\documentclass{article} 
\usepackage{nips12submit_e,times}
\usepackage{graphicx}

\title{A Meta-Theory of Boundary Detection Benchmarks}

\author{
Xiaodi Hou\\
Computation and Neural Systems, Caltech\\
\texttt{xiaodi.hou@gmail.com} \\
\And
Alan Yuille\\
Department of Statistics, UCLA\\
\texttt{yuille@stat.ucla.edu} \\
\And
Christof Koch\\
Computation and Neural Systems, Caltech\\
\texttt{koch@klab.caltech.edu}
}

\nipsfinalcopy

\begin{document}

\maketitle

\begin{abstract}
Human labeled datasets, along with their corresponding evaluation algorithms, play an important role in boundary detection. We here present a psychophysical experiment that addresses the reliability of such benchmarks.  To find better remedies to evaluate the performance of any boundary detection algorithm, we propose a computational framework to remove inappropriate human labels and estimate the intrinsic properties of boundaries.
\end{abstract}

\section{Introduction}
Many problems in human and in computer vision are ill-defined. In problems such as boundary detection, there is no objective measurement that determines whether there is a \emph{perceptually meaningful} boundary in any location in an image. To benchmark the performance of a boundary detection algorithm, human labeled datasets (e.g. BSDS300 \cite{martin2001database} with 200 training images and 100 testing images) play a critical role. These datasets characterize the perceptual definition of boundaries in an implicit way by providing exemplar images that have been labeled by a small number of human subjects.

However, labelers do not always agree with each other. Variability is intrinsically related to the ill-defined nature of boundary detection. Yet there is surprisingly little discussion of data variability for boundary detection and its effect on benchmarks. It is commonly held that the labelers of boundary datasets (such as BSDS300) are reliable.  Examined separately, each boundary seems to be reasonable with some underlying edge in the image. In \cite{martin2001database} Martin \emph{et al.} considers label variability
to be due to different labelers drawing in different levels of details. \cite{martin2001database} believes that even though a labeler may scrutinize some parts of the image in considerable detail, while drawing cursory sketches on other parts, different labelers are consistent in a sense that the dense labels refine the sparse labels without contradicting them. In other words, these different instances of labels all come from the same perceptual hierarchy of an image.

Nevertheless, local consistency within a specific region is not strong enough to legitimatize the entire benchmark. To be able to faithfully evaluate an algorithm, the benchmark data has to be free from both type I (false alarm) and type II (miss) statistical errors. Even though boundaries in a benchmark dataset seem to be reasonable, it is still possible that the labelers may miss some equally important boundaries, leaving us with an imperfect benchmark.  Such benchmark that contains type II errors may incorrectly penalizing an algorithm that detects true boundaries.

We here propose a framework to analyze the quality or \emph{benchability} of any benchmark, and demonstrate with a quantitative experiment that the current dataset for benchmarking can be improved.

\section{Evaluating the risk of a boundary benchmark}\label{sec_risk}
Although different human labels of the same boundary often contain spatial offsets up to several pixels, they rarely contradict each other \cite{martin2001database} (e.g. with one drawing a horizontal and the other a vertical boundary at the same location). Based on these observation, we can merge boundary maps of the same image labeled by different subjects into one master map $\mathbf{y}$. At each pixel location $i$, the response of labeler $l$ is a binary value $y_i^l$ ({\it i.e.}, edge or non-edge). $\mathbf{y}_i = [y_i^1, \ldots, y_i^l \ldots y_i^L]$ concatenates the response of all labelers. We use the assignment algorithm and parameters of \cite{martin2004learning} to determine whether to merge adjacent lines from different subjects at one location.

\begin{figure}[htp]
\centering
\includegraphics[width=13cm]{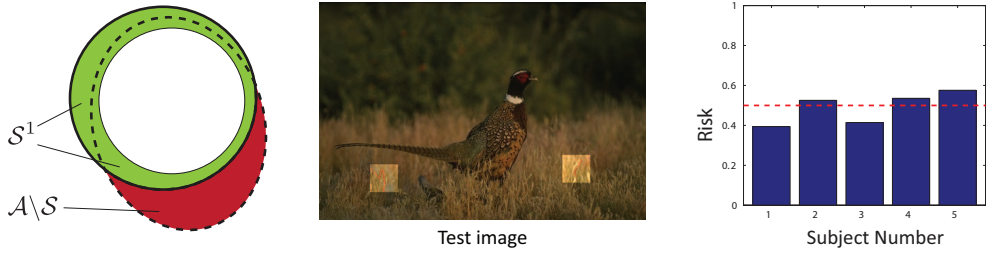}\\
\caption{An illustration of our two-way, forced choice experiment. The left figure shows the Venn diagram of boundary subsets. The thick circle encompasses the full boundary set $\mathcal{S}$.  Within $\mathcal{S}$, the set of orphan labels $\mathcal{S}^1$ is shown in green.  The pB boundary set $\mathcal{A}$ is the dotted ellipsoid. The set of edges falsely identified by the algorithm, $\mathcal{A} \backslash \mathcal{S}$, is highlit in red.  In each trial, we randomly select one boundary segment from $\mathcal{S}^1$ (green ring) and another from $\mathcal{A} \backslash \mathcal{S}$ (red ellipsoid) and ask subjects to judge which one is perceptually stronger. Two boundary segments (high contrast squares with red lines) are superimposed onto the original image (shown in the middle figure).  At the same time, the original is also presented to the subject in a separate window.  In total, $100$ image pairs are compared by all 5 subjects.  The right figure shows the \emph{risk} (that is, how often the false-alarm algorithmic edges are preferred over the human labels), of this database for all 5 subjects. Dotted line is chance level (0.5).}\label{fig_forced_choice}
\end{figure}

To evaluate the correctness of a benchmark, we used a two-way forced choice paradigm (shown in Fig.~\ref{fig_forced_choice}). In any one trial, a subject\footnote{We refer to \emph{labelers} as the people who originally labeled the BSDS300 dataset, while \emph{subjects} refers to people we recruited
that perform our two-way forced choice experiment.} is asked to compare the relative \emph{perceptual strength} of two local boundary segments. Similar to \cite{martin2001database}, we do not give specific instructions that could potentially bias the result towards one particular type of boundary. The advantage of this two-alternative experiment is that it cancels out most of the fluctuations of cognitive factors, such as spatial attention bias, subject fatigue, and decision thresholds that are different in each subject. Moreover, compared to the tedious labeling process, this paradigm is much simpler and cheaper to implemented via crowd-sourcing.

Given sufficient number of comparisons and subjects, we can determine the relative perceptual strength of any pair of boundary segments.  This framework yields a strict total ordering on the set of boundaries.  We can map the boundary set onto the interval $[0, 1]$ by assigning each boundary segment a real-value $x$.  This value $x$ can be considered as the \emph{perceptual strength} of the boundary, because a boundary segment with large $x$, by definition, is stronger (\emph{i.e.}, chosen more frequently by subjects) than another boundary with smaller $x$.  Let $\mathcal{S}$ be the set of all boundaries in a dataset, $s_i$ be one boundary segment from $\mathcal{S}$, and $x_i$ be its perceptual strength. We can define the \emph{risk} of a boundary set $\mathcal{S}$ in relationship to a boundary set $\mathcal{A}$ generated by some reference algorithm as:

\begin{equation}
R(\mathcal{S}, \mathcal{A}) = P(x_i < x_j \mid s_i \in \mathcal{S}, s_j \in \mathcal{A} \backslash \mathcal{S}). \label{eq_risk}
\end{equation}

This paradigm allows us to assess the risk associated with any dataset, such as BSDS300. Because of its great popularity, we choose pB boundaries \cite{martin2004learning} as the reference algorithm set $\mathcal{A}$. We choose the pB threshold such that the number of boundaries in $\mathcal{A}$ is the same as in $\mathcal{S}$ ($\#\mathcal{A} = \#\mathcal{S}$).  To further illustrate the effect, we further restrict the sampling of human labels $s_i$ within a subset we call \emph{orphan labels} $\mathcal{S}^1$, which refers to the boundaries that are labeled by only one labeler ($\mathcal{S}^1 = \{s_i \mid \sum_{l = 1}^L y_i^l= 1\}$) but not by the other $L-1$ labelers \footnote{Sampling human labels from $\mathcal{S}^1$ while algorithm algorithm labels from $\mathcal{A} \backslash \mathcal{S}$ makes the procedure slightly different from the original Eq.~\ref{eq_risk}}. $30.88\%$ of the entire boundary set of BSDS300 are orphan labels.

We used 5 subjects to compare 100 pairs of boundary segments comparison (500 trials in total). For each pair, we use the mode response of all 5 subjects to determine the ordering. The mean risk of $\mathcal{S}^1$ is $0.44$. That is, almost half of the time, a ``false alarm'' algorithmic boundary is perceptually stronger than the orphan label, which would usually be consider ``ground truth''.  Given the large fraction of orphan labels (almost one third of all boundaries), this leaves the validity of using BSDS300 to benchmark any one algorithm in doubt.

Given threshold $\tau$, there exist a \emph{perfect boundary set} $\tilde{\mathcal{S}}_{\tau}$ that has zero risk, such that $x_i \geq \tau$ for any $s_i \in \tilde{\mathcal{S}}_{\tau}$, and $x_j < \tau$ for any $s_j \notin \tilde{\mathcal{S}}_{\tau}$.  This perfect set can be formed by examining boundary strength from all possible boundaries from all images.  However, the current \emph{imperfect} boundary set $\mathcal{S}$ annotated by a finite number of unreliable labelers lacks the information of a vast majority of unlabeled pixels.  There is a probability such that a ``qualified'' boundary $s_i$ with $x_i \geq \tau$ exists in the unlabeled pixels.  This probability decreases as $\tau$ increases, because a relatively strong boundary is less likely to be overlooked by all labelers.  In fact, by taking the extremal threshold $\tau > 1$, we end up with a trivial solution: a risk-free but useless empty boundary set\footnote{The other trivial solution is the original set by setting $\tau$ to $0$.}.

In this paper, we restrict our analysis within existing boundary labels in BSDS300, and try to infer the perceptual strength for each boundary segment.  Inferred perceptual strengthes allow the user to choose an appropriate threshold, and form a subset of boundary segments that balances risk and \emph{utility}, which we refer to the total available number of data-points in the selected subset.  In the next section, we present a graphical model that estimates the boundary perceptual strength.

\section{Model and inference}
\begin{figure}[htp]
\centering
\includegraphics[height=1.5cm]{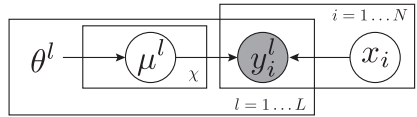}\\
\caption{The graphical model of the labeling process.  This model assumes that the label is determined probabilistically by the perceptual strength $x_i$ and the response profile of the labeler $\mu^l$, which is further controlled by a hidden parameter $\theta^l$.  The gray circle indicates the observed variable, which is the binary individual response to a boundary segment.  The model outputs estimates of the perceptual strength of each boundary segment as well as the parameters of each labeler.}\label{fig_gm}
\end{figure}

During the labeling process, each subject $l$, governed by her/his internal psychophysical parameters $\theta^l$, responds to segments of different perceptual strength $x_i$. For all the boundaries such that $\{i \mid x_i = \chi \}$, the response $y_i^l$ yields a mixture of Bernoulli distributions, with parameter $\mu^l(\chi)$. Furthermore, we assume $\mu^l(\chi)$ yields a sigmoid functional form. The graphical model of the labeling process is shown in Fig.~\ref{fig_gm}.

In our model, $x_i$ yields a uniform distribution $\mathcal{U}(0, 1)$. $\mu^l(\chi) = s(\chi, \theta)$, where $s(\cdot)$ is the sigmoid function: $s(\chi, \theta) = \frac{\theta_3^l}{1 + \exp(\theta_2^l - \theta_1^l \chi)} - \theta_4^l$.  The conditional probability of $y_i^l$ is a soft voting of different $\mu$, such that $P(y_i^l = 1 \mid \mu^l, x_i) = \int_\chi \phi_{\sigma}(x_i - \chi) \mu^l(\chi) \textrm{d}\chi$, where $\phi_{\sigma}(\cdot)$ is the Gaussian probabilistic density function with zero mean and $\sigma$ standard deviation. We set $\sigma = 0.15$.

We use the EM algorithm to estimate $\theta^l, \mu^l(\chi)$, and $x_i$.  We start with $E_l[y_i^l]$ as the initial guess $x_i^{\star}$.  In each iteration, the estimate of $\mu$ is given by $\mu^l(\chi)^{\star} = \sum_i y_i^l \phi_{\sigma}(x_i^{\star} - \chi)$.  $\theta$ is updated by $\theta^{l\star} = \arg\min_{\theta} \int_\chi \big( s(\chi, \theta^l) - \mu^l(\chi) \big)^2 \textrm{d}\chi$.  For the estimate of $x$, we have $x_i^{\star} = \arg\max_{x_i} \prod_l P(y_i^l \mid \mu^l, x_i)$.  The optimization process converges within 20 iterations.  The distribution of the perceptual strength is shown in Fig.~\ref{fig_exp_results}.

\section{Experimental validation}
\begin{figure}[hpb]
\centering
\includegraphics[width=14cm]{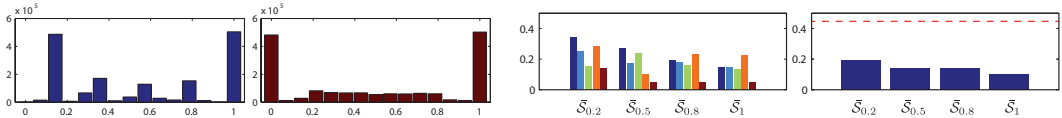}\\
\caption{ Left 1: Initial guess of the perceptual strength distribution.  Left 2: Final estimate of the perceptual strength distribution.  Right 1: Individual estimates of the risk of $\bar{\mathcal{S}}_{\tau_i}$.  In this figure, each color corresponds to one subject.  Right 2: Risk estimate based on majority voting of all subjects.  The dotted line in the right figure indicate the mode risk of $\mathcal{S}_{\tau_i}$ in Fig.~\ref{fig_forced_choice}.}\label{fig_exp_results}
\end{figure}

Given the inferred perceptual strengthes, we select 4 thresholds $\tau_1 = 0.2$, $\tau_2 = 0.5$, $\tau_3 = 0.8$, and $\tau_4 = 1$, and formed 4 subsets $\bar{\mathcal{S}}_{\tau_i}$ of boundary segments.  For each $\bar{\mathcal{S}}_{\tau_i}$ we use the pB algorithm to generate $\mathcal{A}^i$ such that $\#\bar{\mathcal{S}}_{\tau_i} = \# \mathcal{A}^i$.  Finally, a 5-subject experiment is conducted to evaluate the risk of $\bar{\mathcal{S}}_{\tau_i}$ .  For each image, we randomly choose a pair of boundary segments from $\mathcal{A}^i$ and $\bar{\mathcal{S}}_{\tau_i}$, and then take the majority voting of our subjects' responses to estimate the relative strength ordering.  A total number of 500 trials are averaged to estimate the risk of each subset.  The result is shown in Fig.~\ref{fig_exp_results}.

\section{Discussion and future works}

There are two main trends in the perceptual strength distributions shown in Fig.~\ref{fig_exp_results}.  First, the spiky distribution of initial guess has been successfully smoothed out, because each subject has his distinctive labeling characteristics and therefore their response weights differently to the estimated strength.  Second, many of the boundary strengthes are automatically suppressed to zero.  In fact, most of these zero-strength boundary segments correspond to the orphan labels, which are the biggest source of the dataset risk.  From the right two figures, we see that the subset risk decreases as the perceptual strength threshold $\tau$ goes up.  This result supports the risk-utility model we mentioned in Sec.~\ref{sec_risk}.

We have shown that a human-labeled dataset, even if well constructed and tested, can contain serious risks that hinder its ability to evaluate algorithm performance. We first proposed a psychophysical test to estimate human dataset risk, where by risk we mean mistakenly classifying strong algorithmic boundaries as false alarms.  We discuss an inference model to find the perceptual strength of each boundary segment, and use it to balance the risk utility trade-off.

Due to space limitation, we are unable to discuss other factors such as the stability of labeler-image assignment and its influence on the perceptual strength estimation; the information-theoretic limit of the two-way force choice, and result variation by using different algorithms.  These issues will be addressed in the journal submission of this paper \cite{hou2012meta}.

\subsubsection*{Acknowledgments}
\small
The first author would like to thank Liwei Wang, Yin Li, Xi (Stephen) Chen, and Katrina Ligett.  The research was supported by the ONR via an award made through Johns Hopkins University and by the Mathers Foundation.

\bibliographystyle{plain}
\small{
\bibliography{myBib}

\begin{thebibliography}{1}

\bibitem{hou2012meta}
X.~Hou, C.~Koch, and A.~Yuille.
\newblock A meta-theory of boundary detection benchmarks.
\newblock {\em in preparation}.

\bibitem{martin2001database}
D.~Martin, C.~Fowlkes, D.~Tal, and J.~Malik.
\newblock A database of human segmented natural images and its application to
  evaluating segmentation algorithms and measuring ecological statistics.
\newblock In {\em Computer Vision, 2001. ICCV 2001. Proceedings. Eighth IEEE
  International Conference on}, volume~2, pages 416--423. IEEE, 2001.

\bibitem{martin2004learning}
D.R. Martin, C.C. Fowlkes, and J.~Malik.
\newblock Learning to detect natural image boundaries using local brightness,
  color, and texture cues.
\newblock {\em Pattern Analysis and Machine Intelligence, IEEE Transactions
  on}, 26(5):530--549, 2004.

\end{thebibliography}
}

\end{document}